\documentclass[10pt,twocolumn,letterpaper]{article}

\usepackage[pagenumbers]{wacv} 

\usepackage{graphicx}
\usepackage{amsmath}
\usepackage{amssymb}
\usepackage{booktabs}
\usepackage{multirow}
\usepackage{pifont} 

\usepackage[pagebackref,breaklinks,colorlinks]{hyperref}

\usepackage[capitalize]{cleveref}
\crefname{section}{Sec.}{Secs.}
\Crefname{section}{Section}{Sections}
\Crefname{table}{Table}{Tables}
\crefname{table}{Tab.}{Tabs.}
\usepackage{xcolor}

\def\wacvPaperID{1593} 
\def\confName{WACV}
\def\confYear{2025}

\begin{document}

\title{Multimodal Fusion Learning with Dual Attention for Medical Imaging}

\author{
    Joy Dhar\textsuperscript{1} \and 
    Nayyar Zaidi\textsuperscript{2} \and 
    Maryam Haghighat\textsuperscript{3} \and 
    Puneet Goyal\textsuperscript{1,6} \and 
    Sudipta Roy\textsuperscript{4} \and 
    Azadeh Alavi\textsuperscript{5} \and Vikas Kumar\textsuperscript{1} \and \\
    {\small \textsuperscript{1}Indian Institute of Technology Ropar} \quad 
    {\small \textsuperscript{2}Deakin University} \quad 
    {\small\textsuperscript{3}Queensland University of Technology} \quad 
    \small\textsuperscript{4}Jio Institute, India \quad 
     \\ {\small \textsuperscript{5} RMIT University} \quad
      {\small\textsuperscript{6}NIMS University, Jaipur, India} 
       }

\maketitle
\begin{abstract}
Multimodal fusion learning has shown significant promise in classifying various diseases such as skin cancer and brain tumors. However, existing methods face three key limitations. First, they often lack generalizability to other diagnosis tasks due to their focus on a particular disease. 
Second, they do not fully leverage multiple health records from diverse modalities to learn robust complementary information. 
And finally, they typically rely on a single attention mechanism, missing the benefits of multiple attention strategies within and across various modalities. 
To address these issues, this paper proposes~\emph{a dual robust information fusion attention mechanism}~\texttt{(DRIFA)} that leverages two attention modules -- i.e., multi-branch fusion attention module and the multimodal information fusion attention module. 
\texttt{DRIFA} can be integrated with any deep neural network, forming a  multimodal fusion learning framework denoted as~\texttt{DRIFA-Net}. 
We show that the multi-branch fusion attention of~\texttt{DRIFA} learns enhanced representations for each modality, such as dermoscopy, pap smear, \texttt{MRI}, and \texttt{CT-scan}, whereas multimodal information fusion attention module learns more refined multimodal shared representations -- improving the network's generalization across multiple tasks and enhancing overall performance. 
Additionally, to estimate the uncertainty of~\texttt{DRIFA-Net} predictions, we have employed an ensemble Monte Carlo dropout strategy. 
Extensive experiments on five publicly available datasets with diverse modalities demonstrate that our approach consistently outperforms state-of-the-art methods. The code is available at \url{https://github.com/misti1203/DRIFA-Net}.
\end{abstract}

\section{Introduction} \label{sec:intro}

Recent advancements in machine learning (\texttt{ML}) for medical imaging analysis, particularly in cancer classification, have transformed healthcare practices  enabling quick and cost-effective decision-making for physicians and potentially saving lives \cite{Dhar2021a}.
It is important to note that there are various imaging modalities prevalent in medical domain such as dermoscopy, pap smear, \texttt{MRI}, and \texttt{CT scans} -- and are crucial for detecting cancers like skin, cervical, brain tumors, and lung cancer. 
Existing methods for handling multiple modalities relies on building models on single modality and then leverage techniques such as transfer learning (\texttt{TL}), feature fusion, attention mechanisms, etc. to exploit knowledge of each model~\cite{steyaert2023multimodal, celik2024development, Hemalatha2023, Anand2023, Han2024, Abdelhalim2021, Ling2023, Yutra2023, Yang2023, Qian2022, Kim2023, Naveed2024, Zhou2021, Pedro2022}.
However, reliance on single-modal learning approaches often results in sub-optimal performance due to inefficient feature extraction and noise in the data, leading to over-fitting as well. 
How to learn an effective model that can leverage various modalities -- also known as~\emph{Multimodal fusion learning} (\texttt{MFL}), has been an open question in machine learning.
Multimodal fusion learning integrates information from multiple modalities to enhance representation and therefore improve predictive performance \cite{islam2022mumu}.
It aims to address various challenges faced by single-modal models by learning shared representations from diverse modalities.

In the last few years, attention-based models have rose to popularity which automatically learn the importance of any individual token (i.e., element of interest)~\cite{vaswani2017attention}, and multimodal fusion learning has not been any exception. 

E.g.,~\cite{huang2021gloria, cheng2022fully, georgescu2023multimodal, Cai2023, He2023, Omeroglu2023} are some notable attention-based works that have been developed to learn robust representations from medical imaging modalities.
However, these approaches face significant challenges. Firstly, they often have limited capacity to learn shared complementary information leading to sub-optimal performance.  
Also, their focus on specific a modality such as \texttt{MRI}, \texttt{PET}, and \texttt{SPECT} for brain disorders or dermoscopy for skin cancer restricts their generalizability. 
Secondly, they typically rely on single attention mechanisms, hence do not avail the opportunity to utilize multiple attention strategies to independently enhance multimodal representation learning across different modalities. 
These limitations underscore the need for more robust methodology capable of addressing these issues.

To address these challenges, we propose a dual robust information fusion attention mechanism integrated within a deep neural network, denoted as~\texttt{DRIFA-Net}. Our proposed multimodal fusion learning strategy incorporates two attention mechanisms: a) multi-branch fusion attention enhances representations within each modality, and b) multimodal information fusion attention enhances multimodal representations to improve our learning model's performance. 
 
In summary, our main contributions are as follows:
\begin{itemize}
    \item[$\bullet$] We propose a dual robust information fusion attention mechanism to enhance~\texttt{MFL} denoted as~\texttt{DRIFA-Net}.
    \item[$\bullet$] We design a multi-branch fusion attention module (\texttt{MFA}) that combines the strengths of hierarchical information fusion attention and channel-wise local information attention modules to efficiently learn diverse local dependencies. 
    \item[$\bullet$] We devise a multimodal information fusion attention (\texttt{MIFA}) module that incorporates global and local information fusion attention modules to effectively learn multimodal global-local dependencies.
    \item[$\bullet$] We conduct comprehensive comparisons with prior state-of-the-art approaches to demonstrate the effectiveness of our attention method on diverse medical imaging datasets: \texttt{HAM10000} \cite{tschandl2018ham10000}, \texttt{SIPaKMeD} \cite{plissiti2018sipakmed}, \texttt{NickParvar} \cite{nickparvar2021}, \texttt{Lung CT-scan} \cite{alyasriy2020iq}, and \texttt{BraTS2020} \cite{menze2014multimodal}. 
    \item[$\bullet$] Finally, we employ the ensemble Monte Carlo dropout strategy to estimate uncertainty in our \texttt{DRIFA-Net}'s predictions.
\end{itemize}
The rest of the paper is organized as follows. We discuss related works in Section~\ref{sec:rel} followed by the proposition of our proposed method in Section~\ref{sec:method}.
The experimental analysis is conducted in Section~\ref{sec:exp}. We conclude in Section~\ref{sec:concl} with pointers to future works.

\begin{figure*}[ht!]
    \centering
    \includegraphics[width=0.95\textwidth, height=0.2\textheight]{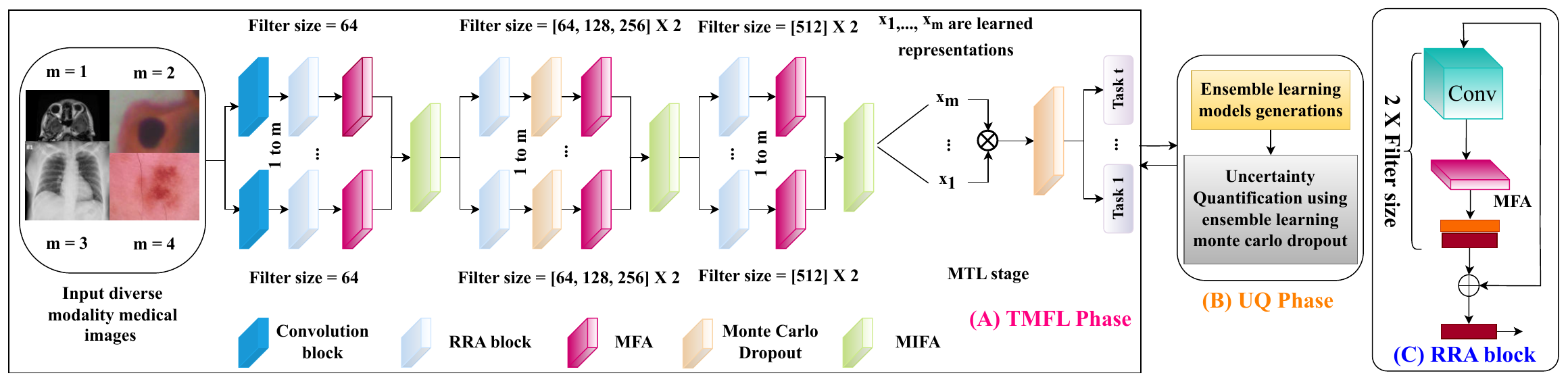}

    \caption{Detailed architecture of~\texttt{DRIFA-Net}. Key components include: \textbf{(A)} the target-specific multimodal fusion learning (\texttt{TMFL}) phase, followed by \textbf{(B)} an uncertainty quantification (\texttt{UQ}) phase.
    \texttt{TMFL} phase comprises a robust residual attention (\texttt{RRA}) block, shown in \textbf{(C)}, and utilizes multi-branch fusion attention (\texttt{MFA}), an additional \texttt{MFA} module for further refinement of local representations, a multimodal information fusion attention (\texttt{MIFA}) module for improved multimodal representation learning, and multitask learning (\texttt{MTL}) for handling multiple classification tasks. During (\texttt{UQ}) phase,  the reliability of~\texttt{DRIFA-Net} predictions are assessed.}
    \label{fig:fig2}
\end{figure*}

\section{Related Works} \label{sec:rel}

Prior studies use various attention mechanisms with single-modal learning approaches to detect across different medical imaging modalities.
E.g., in dermoscopy modality, different attention strategies \cite{Zhou2021, Pedro2022, Qian2022, Yang2023, Kim2023, Yutra2023, Anand2023, Ling2023, Naveed2024, Han2024} have been employed  to learn fine-grained details for skin cancer classification and segmentation tasks. 
For \texttt{MRI} modality,~\cite{alzahrani2023convattenmixer} used self-attention strategies. 
For pap smear and \texttt{CT-scan} modalities, existing works \cite{Hemalatha2023, Manna2021, Song2024, Liu2022, Pacal2023} primarily focused on feature fusion or \texttt{TL} strategies rather than attention mechanisms.

Multimodal fusion learning addresses the aforementioned limitations by enhancing modality strengths and thereby improving classification \cite{islam2022mumu}. 

For instance, Chen et al. \cite{Chen2023JCRCO} developed a multimodal data fusion diagnosis network \texttt{(MDFNet)} for skin cancer classification integrating clinical images and patient data. 
Li et al. \cite{li2021multimodal} designed a multimodal medical image fusion technique by decomposing medical images to capture rich gradients, benefiting \texttt{MRI} analysis. 
Kihara et al. \cite{kihara2022policy} used a hybrid deep learning method combining paired medical images for enhanced clinical predictions. 
Tan et al. \cite{tan2022multi} introduced the multi-CoFusion approach, applying multimodal fusion learning for glioma grade classification and survival analysis using histopathological and mRNA data. 
Tabarestani et al. \cite{tabarestani2020distributed} deployed a distributed~\texttt{MFL} approach for Alzheimer’s disease prediction using \texttt{MRI} and \texttt{PET} modalities.

In the following let us discuss attention-based approaches to~\texttt{MFL}.
Huang et al. \cite{huang2021gloria} introduced a label-efficient multimodal medical imaging representation method, incorporating radiology reports and a global-local attention mechanism to learn global-local representations. 
Cheng et al. \cite{cheng2022fully} proposed an \texttt{MRI}-based multimodal fusion learning approach using a hybrid \texttt{CNN-Transformer} for Glioma Segmentation and a multi-scale classifier for \texttt{IDH} genotyping on the \texttt{BraTS2020} dataset. 
Georgescu et al. \cite{georgescu2023multimodal} devised a multimodal multi-head convolutional attention module for super-resolving \texttt{CT} and \texttt{MRI} scans. 
Cai et al. \cite{Cai2023} developed a multimodal transformer with separate encoders for images and metadata fusion, using a vision transformer \texttt{(ViT)} and a mutual attention \texttt{(MA)} block to enhance feature fusion on \texttt{HAM10000} \cite{tschandl2018ham10000}. 
He et al. \cite{He2023} proposed a co-attention fusion \texttt{(CAF)} network for multimodal skin cancer diagnosis on the seven-point checklist dataset \cite{kawahara2018seven}, utilizing co-attention \texttt{(CA)} and attention fusion \texttt{(AF)} blocks. 
Omeroglu et al. \cite{Omeroglu2023} developed a soft attention-based \texttt{MFL} network for multi-label skin lesion classification, leveraging multiple branches to learn complementary features.

\section{Proposed Method} \label{sec:method}

In this section, we will introduce our proposed model~\texttt{DRIFA-Net}.  The input features from \( m \) heterogeneous modalities are represented as \( X = [x_1, x_2, \ldots, x_m] \), and their corresponding labels \( [y_1, \ldots, y_t] \) -- the model is expected to perform \( t \) (binary or multi-class) classification tasks. \( X \) is used as input to~\texttt{DRIFA-Net} \( \theta(\cdot) \) to learn an enhanced multimodal shared representations denoted as: \( X^S = \theta(X) = [x_1^s, x_2^s, \ldots, x_m^s] \). Here, $x_1^s, x_2^s, \ldots, x_m^s$ denote the enhanced learned representations obtained from each branch of~\texttt{DRIFA-Net}. 
In this study, we address specific target tasks \( y_t \in [0, 1, \ldots, n] \), where n denotes the number of classes, involving both binary and multi-class classification problems across the diverse modalities present in our datasets.

\subsection{Method Overview} \label{sub:overview}

Here we delve into the details of our proposed dual robust information fusion attention mechanism~\texttt{DRIFA}. 

We will show that it leverages information fusion learning with attention strategies across multimodal fusion learning settings, ensuring versatility and adaptability within diverse neural architectures.
\texttt{DRIFA} is integrated with~\texttt{ResNet18}, creating a multi-branch multimodal fusion learning network denoted as~\texttt{DRIFA-Net} (depicted in Fig.~\ref{fig:fig2}). 
\texttt{DRIFA} method is detailed in Algorithm \textcolor{green}{1}, which can be found in the supplementary materials. 
Each branch within~\texttt{ResNet18} incorporates input features from the corresponding modality, where \texttt{ResNet18} structure consists of one convolutional block and eight residual blocks tailored to learn representations for each modality. 
In the following, we will discuss two salient phases of~\texttt{DRIFA-Net}, i.e., -- target-specific multimodal fusion learning (\texttt{TMFL}) and uncertainty quantification (\texttt{UQ}). 
\begin{figure}[t]
    \centering
\includegraphics[width=0.479\textwidth, height=0.33\textheight]{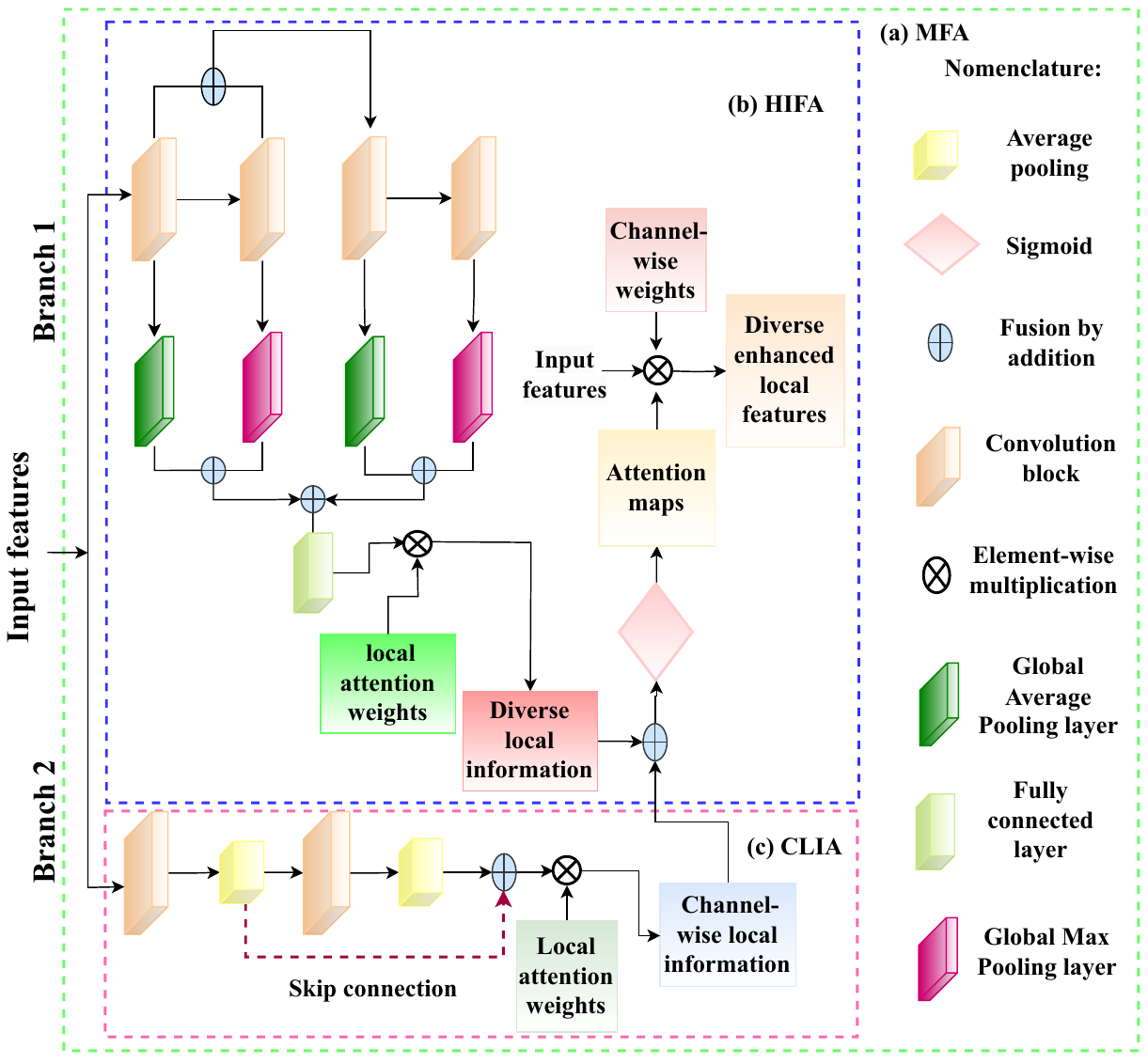}
    \caption{(a) Multi-branch fusion attention (\texttt{MFA}) module. Key components include hierarchical information fusion attention (\texttt{HIFA}) for diverse local information enhancement and channel-wise local information attention (\texttt{CLIA}) for improved channel-specific representation learning.}
    \vspace{-.5cm}
    \label{fig:fig3}
\end{figure}

\subsection{\textbf{Target-specific Multimodal Fusion Learning}} \label{sub:tmfl}

\texttt{DRIFA-Net} relies on target-specific multimodal fusion~(\texttt{TMFL}), in pursuit of learning an enhanced shared multimodal representations to achieve better performance across target specific classification tasks. 
\texttt{TFML} utilizes a robust residual attention (\texttt{RRA}) block, which incorporates our proposed multi-branch fusion attention (\texttt{MFA}) module, which effectively learns diverse refined local patterns. 
Additionally, \texttt{TMFL} also incorporates our proposed multimodal information fusion attention (\texttt{MIFA}) module to learn enhanced multimodal representations.
Finally, a target-specific multitask learning (\texttt{MTL}) approach is used to handle multiple classification tasks simultaneously within \texttt{TMFL} phase.
In the following, we will discuss~\texttt{RRA},~\texttt{MIFA} and~\texttt{MTL} blocks -- which represents salient elements of interest in~\texttt{TMFL}.

\subsubsection{RRA: Robust Residual Attention block} \label{subsubsec_RRA}
Let us in this section discuss~\texttt{RRA} block which incorporates our proposed \texttt{MFA} module applied after each convolutional layer and utilizes a skip connection strategy. 
This approach aims to learn diverse local representations, thereby enhancing the performance of our learning network.

\textbf{Multi-branch Fusion Attention Module~(\texttt{MFA})} module aims to learn enhanced local representations from input features. It is illustrated in Fig.~\ref{fig:fig3}.
Specifically, the \texttt{MFA} module is integrated within each~\texttt{RRA} block across all branches of the network. 
Another~\texttt{MFA} module is employed to further refine these representations (Fig.~\ref{fig:fig2}~(a)), thereby improving the model's ability to learn more detailed local patterns. 

To enhance local information acquisition, \texttt{MFA} utilizes two attention modules: 
\begin{itemize}
    \item Hierarchical Information Fusion Attention (\texttt{HIFA}) module which enriches diverse local information, and 
    \item Channel-Wise Local Information Attention (\texttt{CLIA}) module which compresses channel-wise information. 
\end{itemize}
\texttt{HIFA} module is integrated into the first branch to capture diverse local features, while the~\texttt{CLIA} module is applied in the second branch to refine channel-wise information, as shown in Figure \ref{fig:fig3}. 
Additionally, a modulation strategy is employed to selectively emphasize critical representations in the input data and suppress irrelevant ones, thereby enhancing the overall performance of our learning network.

The~\texttt{MFA}  module aims to enhance diverse local representation learning by transforming input feature maps $x \in \mathbb{R}^{H \times W \times C}$, where $H$, $W$, and $C$ denote the height, width, and number of channels respectively, to 
$x^{\prime} = x \otimes a \otimes \omega_{c}$. \label{eq}
Here, $\otimes$ denotes element-wise multiplication, $a$ is enhanced local attention maps, and $\omega_{c}$ is channel-wise learnable parameters  that adjust the importance of each channel during training.

To design the~\texttt{HIFA} module, we use $p$-th 1x1 convolution layers denoted as $\psi_{p}$, $\frac{p}{2}$-th global average pooling (\texttt{GAP}) layers denoted as $\beta$, and $\frac{p}{2}$-th global max pooling (\texttt{GMP}) layers denoted as $\gamma$ for learning diverse local information. 
The process involves four key steps: 
\begin{itemize}
    \item First, input features are processed through a convolution layer and a \texttt{GAP} layer to capture initial local information \( l_{p=0} \). 
    \item Secondly, features from the \( p \)-th convolution layer are refined using a subsequent convolution layer and a \texttt{GMP} layer to extract additional local information \( l_{p=1} \). 
    \item Thirdly, the refined features are fused and passed through further convolution layers, each followed by either \texttt{GAP} or~\texttt{GMP}, to capture diverse local information \( l_{p} \). 
    \item Finally, the resulting local information variants are hierarchically fused to obtain enhanced diverse local patterns. E.g., local information \( l_{p=0} \) is fused with other local patterns \( l_{p, p \neq 0 }\) 
    to learn enhanced diverse local information, followed by application of a fully connected layer \( f \) to compress these enhanced diverse information $\hat{d}$, as illustrated in Figure~\ref{fig:fig3}. This  can be written as:

    \begin{eqnarray} \label{eq:eq2}
    l_{p} & = & \forall_{p} (\beta | \gamma) \circ (\Pi_{p}, \lambda_1, \lambda_2, \lambda_3), \textrm{ where } \nonumber \\
    \lambda_1 & = & \psi_{(p+1)}(\Pi_p), \nonumber \\
    \lambda_2 & = & \psi_{(p+2)}(\phi(\Pi_p, \lambda_1)), \nonumber \\
    \lambda_3 & = & \psi_{(p+3)}(\lambda_2). \nonumber \\
    \end{eqnarray}

\vspace{-1mm}
Here, \(\phi\) represents the fusion (addition) strategy, \( \forall_p \) denotes ``for all \( p \)'', 
\( \Pi_p \) represents \( \psi_p(x) \), \( \circ \) denotes composition operator, and 
    \( | \) indicates either to employ the \( \beta \) layer or the \( \gamma \) layer after processing each components (as process), such as \( \Pi_p \), $\lambda_1$, $\lambda_2$, and $\lambda_3$ to learn $l_p$. 

    \vspace{-3mm}
    \begin{equation} \label{eq:eq3}
    \hat{d} = f(\forall_{l_p} \left[ \varphi \{ H_i, H_{i+1} \} \right]),
    \end{equation}
    where $\varphi$ denotes fusion (concatenation), $\hat{d}$ represents diverse enhanced local information for each $m$, $ H_i = \varphi(l_p, l_{p+1}) $ and $ H_{i+1} = \varphi(l_{p+2}, l_{p+3}) $ such that $ i = 0, 1 $.

\end{itemize}

To design our~\texttt{CLIA} module, we utilize \( q \)-th 1x1 convolution layers with sigmoid activation function ($\sigma$) followed by \( q \)th average pooling layer to compress channel information. 
Additionally, we use a skip connection strategy to fuse the resulting information with the initially compressed channel information, enhancing the learning of channel-wise local information and thereby improving our model's performance, as we have:

\begin{equation} \label{eq:eq4}
\hat{l} = \phi(\delta_{q=1}(\eta_1), \delta_{q=2}(\eta_2))
\end{equation}

where $\delta$ denotes average pooling layer, $\hat{l}$ represents enhanced local information i.e.,
\vspace{-1mm}
\begin{eqnarray}
\eta_1 & \in & \sigma_{(q=1)}(\psi_{(q=1)}(x)), \nonumber \\
\eta_2 & \in & \sigma_{(q=2)}(\psi_{(q=2)}(\delta_{q=1}(\eta_1))). \nonumber
\end{eqnarray}

Finally, to combine the local information 
learned from ~\texttt{HIFA} and~\texttt{CLIA},  we use learnable weights \(\omega_d\) and \(\omega_l\) to adjust the importance of each learned local information component that is \(\hat{d}\) and \(\hat{l}\).
Initially, \(\omega_d\) is set to one and is multiplied with diverse enhanced local information \(\hat{d}\) to refine these patterns. 
Similarly, \(\omega_l\) is set to one and multiplied with channel-wise enhanced local patterns \(\hat{l}\) to focus on refining channel-specific information. 
The refined information is then fused to enhance the capture of diverse local details. 
A sigmoid activation function \(\sigma\) is applied to generate attention maps \(a\), which highlight crucial features and improve network performance by capturing fine-grained details, as shown in the following equation:
\begin{equation} \label{eq:eq5}
    a = \sigma ((\hat{d} \otimes \omega_d) + (\hat{l} \otimes \omega_l)).
\end{equation}

These learnable weights in the \texttt{MFA} module are activated using a boolean value and are optimized through a back-propagation strategy, as detailed in the supplementary materials.

\subsubsection{\textbf{MIFA: Multimodal Information Fusion Attention Module}} \label{subsubsec_MIFA}

\begin{figure}[t]
    \centering
    \includegraphics[width=0.48\textwidth, height=0.38\textheight]{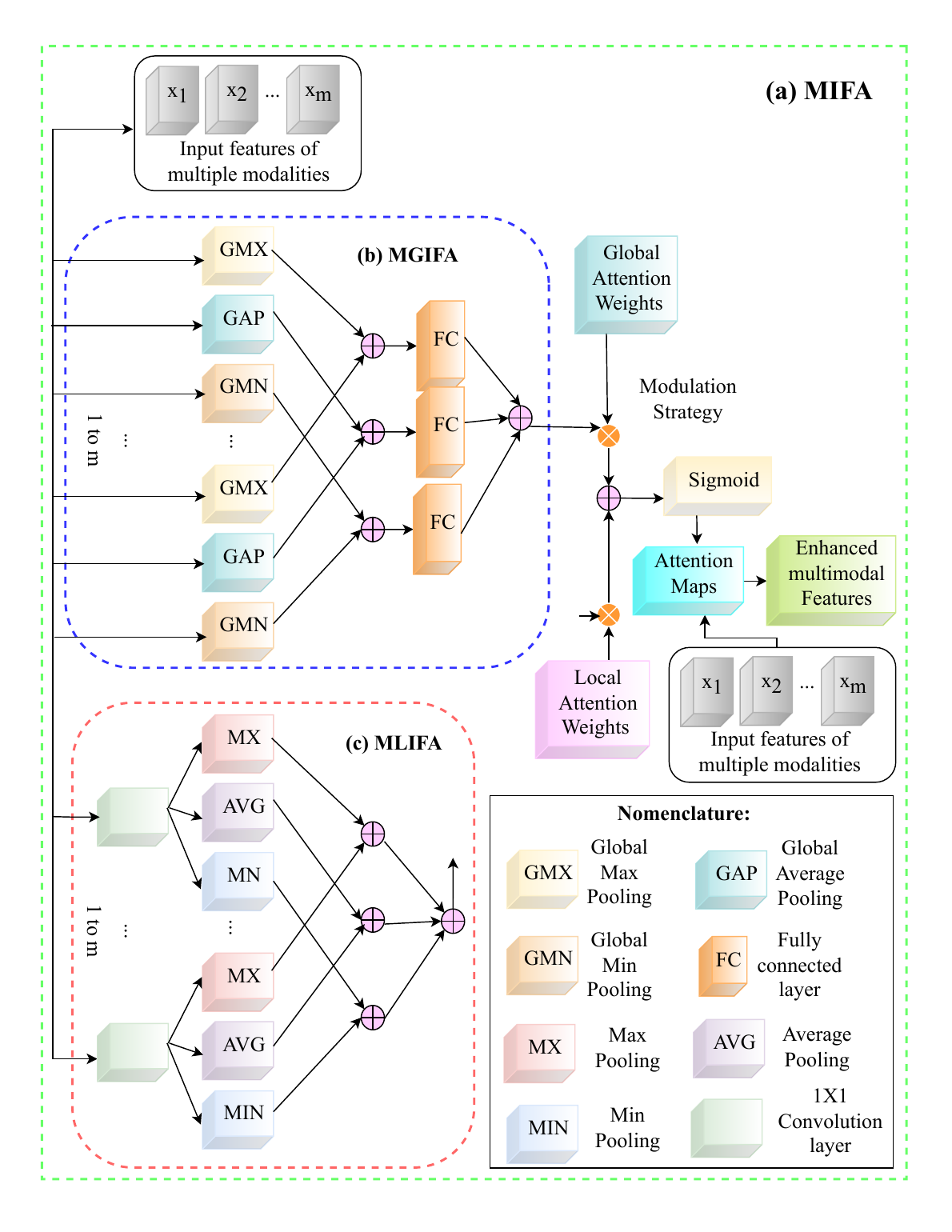}
    \caption{(a) Multimodal information fusion attention (\texttt{MIFA}) module. This module includes multimodal global information fusion attention (\texttt{MGIFA}) (shown in \textbf{b}) and multimodal local information fusion attention (\texttt{MLIFA}) (shown in \textbf{c}).}
    \label{fig:fig4}
\end{figure}
Let us discuss our proposed~\texttt{MIFA} module in this section.
The module is illustrated in Figure~\ref{fig:fig4}.

Given input feature maps from \( m \) heterogeneous modalities, \( X \in x'_m \) (where \( x'_m \) denotes the enhanced local representations for each modality), the~\texttt{MIFA} module aims to learn enhanced multimodal representations \( X^S \). 

This involves element-wise multiplication of multimodal shared attention maps \( A \) with input features \( X \), and channel-wise learnable parameters \( \omega_{c_m} \) for each $m$, i.e., 
\vspace{-1mm}
\begin{equation} \label{eq:eq6}
X^S = X \otimes A \otimes w_{c_m}.
\vspace{-.1cm}
\end{equation}
These parameters work similarly to the channel-wise learnable parameters of the~\texttt{MFA} module (subsection \ref{eq}), i.e., $w_{c_m}$ that adjust the importance of each channel during training. 
However, unlike~\texttt{MFA}, which is applied to a single modality, these parameters are utilized across multiple modalities.

To learn multimodal shared attention feature maps \( A \), we design two attention modules: the multimodal global information fusion attention (\texttt{MGIFA}) and the multimodal local information fusion attention (\texttt{MLIFA}). 
Additionally, we incorporate a fusion strategy similar to the~\texttt{MFA} module. 

Both~\texttt{MGIFA} and~\texttt{MLIFA} modules incorporate various pooling layers. 
For learning diverse global contexts -- global minimum pooling ($\alpha$), global max pooling ($\gamma$), and global average pooling ($\beta$) are used.
Whereas for learning diverse local fine-grained details -- minimum pooling $\vartheta$, 
max pooling $\tau$, 
and average pooling $\delta$ 
are used.

To enhance the learning of diverse global and local information, we design the multimodal global and local information fusion~\texttt{(MGLIF)} approach. 
Specifically, this approach fuses each pooling layer of one modality with corresponding similar pooling layers of other modalities to learn complementary information in both global and local contexts. 
The resulting complementary information enhances learning across all modalities in each branch of our learning network model, bolstering a better performance. 
For example, to enhance global information obtained from global average pooling, the learned global information from modality $1$ is fused with the global information from the other \( m-1 \) modalities. 
This strategy is applied uniformly across all information learned from respective pooling layers, aiming to achieve enhanced diversity in both global and local information. Furthermore, a fully connected layer $f_{pool}$  is applied to each resulting information, followed by the fusion of all resulting information to learn enriched global $g'$ and local representations $l'$ -- as shown in the following equation:

\vspace{-1mm}
\begin{equation} \label{eq:eq7}
g' = \phi \left( f_{pool} \left( \sum_{i=1}^{m} G_{pool,i} (X) \right) \right)
\end{equation}

\begin{equation} \label{eq:eq8}
l' = \phi \left( f_{pool} \left( \sum_{i=1}^{m} L_{pool,i} (X) \right) \right)
\end{equation}

where \( X \in x'_m \), \( G_{pool} \in \{\alpha, \gamma, \beta\} \) and \( L_{pool} \in \{\vartheta, \tau, \delta \}  \).

Similar to the \texttt{MFA} module, to combine learned information (e.g., \(\hat{g}\) and \(\hat{l}\)) from \texttt{MGIFA} and \texttt{MLIFA}, we use learnable weights (\(\omega_{d_m}\) and \(\omega_{l_m}\)) to adjust the importance of this information, thereby refining the patterns. In multimodal fusion learning settings, a fusion operation (addition) followed by a sigmoid activation \(\sigma\) generates multimodal shared attention maps \(A\), capturing diverse global contexts and fine-grained details:
\vspace{-1mm}
\begin{equation}
    \label{eq:mifa_modulation}
    A = \sigma ((g' \otimes \omega_{d_m}) + (l' \otimes \omega_{l_m})).
    \vspace{-2mm}
\end{equation}

\subsubsection{\textbf{MTL: Target-specific Multitask Learning}}

In the~\texttt{MTL} stage, we utilize shared representations \( X^S \) from the~\texttt{TMFL} phase across \( m \) diverse medical imaging modalities. This enhances \texttt{DRIFA-Net}'s generalization capability by learning robust complementary information, thereby improving predictions on multiple modality-specific test sets.
The~\texttt{MTL} approach leverages~\texttt{DRIFA-Net} \( \theta(\cdot) \) to map input features \( [x_1, \ldots, x_m] \) from \( m \) modalities to \( t \) classification tasks \( [y_1, \ldots, y_t] \). 
The~\texttt{MTL} loss function \( \partial_{\text{MTL}} \) combines task-specific cross-entropy losses \( \partial_t^m \), as defined in the following as:
\vspace{-2mm}
\begin{equation} \label{eq:eq10}
    \partial_{\texttt{MTL}} = \sum_{t} \omega_t^m \times \partial_t^m (\theta(X^S, y_t)).
\vspace{-.3cm}
\end{equation}

where \(\theta(X^S, y_t) = [x_1, \ldots, x_m] \rightarrow [y_1, \ldots, y_t]\), and \(\omega_t^m\) represents the weighting factor for each task-specific cross-entropy loss, ensuring efficient task performance balance.

\subsection{Uncertainty Quantization}

We assess the prediction uncertainty in~\texttt{DRIFA-Net} using the ensemble Monte Carlo dropout strategy. 
This approach computes soft probabilities \( \hat{y} \) by averaging random predictions from \( z \) ensemble models, each utilizing stochastic sampling dropout masks \( (\aleph) \) to introduce randomness into~\texttt{DRIFA-Net}. 
Our approach involves \( e = 20 \) iterations of Monte Carlo sampling to generate diverse predictions through \( \theta(\cdot) \). 
During testing,~\texttt{DRIFA-Net} is executed \( 20 \) times on multiple modality-specific test sets, and prediction uncertainty is inferred from the averaged results of these runs as shown in the following equation:
\vspace{-2.2mm}
\begin{equation}
\hat{y} = \arg \max \left(\frac{1}{z} \sum_{0}^{z-1} \left[\Omega(\theta(X^S, \aleph))\right]\right),
\end{equation}
where \( \Omega \) represents a softmax classifier.

\section{Experiments and Results} \label{sec:exp}

\textbf{Datasets --} Our experiments utilized five medical imaging datasets: \texttt{HAM10000}~\cite{tschandl2018ham10000}, \texttt{SIPaKMeD} \cite{plissiti2018sipakmed}, \texttt{Nickparvar MRI} \cite{nickparvar2021}, \texttt{IQ-OTHNCCD lung cancer} \cite{alyasriy2020iq}, and \texttt{BraTS2020} \cite{menze2014multimodal} (denoted as~\texttt{D1, D2, D3, D4, and D5} respectively). 
These datasets encompass diverse modalities: dermoscopy, single-cell pap smear, \texttt{MRI}, and \texttt{CT-scan}. 
\texttt{HAM10000} comprises $10,015$ images across seven classes, \texttt{SIPaKMeD} includes $4,049$ images over five classes, \texttt{Nickparvar} contains $7,023$ \texttt{MRI} images across four classes, and the \texttt{IQ-OTHNCCD lung cancer} dataset has $1,098$ images in three classes. 
The \texttt{BraTS2020} dataset comprises four \texttt{MRI} modalities: fluid-attenuated inversion recovery \texttt{(FLAIR)}, T1-weighted \texttt{(T1)}, T1-weighted contrast-enhanced \texttt{(T1ce)}, and T2-weighted \texttt{(T2)} imaging. It includes 369 training subjects, which are randomly split into 201 for training, 35 for validation, and 133 for testing. The \texttt{BraTS2020} challenge provides ground truth annotations for the imaging data, evaluating three sub-regions: whole tumor \texttt{(WT)}, tumor core \texttt{(TC)}, and enhancing tumor \texttt{(ET)}.
Data augmentation techniques, such as rotation and transformation, ensured uniformity in sample size for multimodal fusion learning operations. 
All images were resized to $128 \times 128 \times 3$ pixels, with an $80\%$ training, $10\%$ validation, and $10\%$ testing split. 

\textbf{Models --} 
We compare the performance of \texttt{DRIFA-Net} with state-of-the-art (\texttt{SOTA}) methods across four datasets (D1–D4) for classifying skin cancer, cervical cancer, brain tumors, and lung cancer, as well as one dataset \texttt{(D5)} for brain tumor segmentation tasks.
Specifically, we employ approaches such as \texttt{Gloria} \cite{huang2021gloria}, \texttt{MTTU-Net} \cite{cheng2022fully}, \texttt{CAF} \cite{He2023}, and \texttt{MTF with MA} \cite{Cai2023} (denoted as \texttt{M1, M2, M3, M4} respectively). 
Note,~\texttt{DRIFA-Net} is denoted as \texttt{M5}. 
We reconfigure these methods according to our multimodal fusion learning settings to ensure an accurate and consistent performance comparison.

\textbf{Notation --} In our results, Acc denotes accuracy, Prec represents precision, Rec stands for recall, F1 refers to the F1 score, SN indicates sensitivity, and SP represents specificity. 

\textbf{Implementations Details --} Implementations were executed on an \texttt{NVIDIA GeForce RTX 4060 Ti GPU}. Models were trained for $200$ epochs using cross-entropy loss with a batch size of $32$. 
The Adam optimizer was used with an initial learning rate of $0.001$. 
A Reduce Learning Rate on Plateau scheduler was employed, reducing the rate by a factor of $0.2$ after $5$ epochs of no improvement, with a minimum learning rate of $10^{-5}$. 
For uncertainty quantification, we adopted the ensemble Monte Carlo dropout method, generating five ensemble models with a dropout rate of $0.25$. 
In this study, all learnable parameters, including \(\omega_{c}\), \(\omega_{d}\), \(\omega_{l}\), \(\omega_{d_m}\), \(\omega_{l_m}\), and \(\omega_{c_m}\), are initialized to $1$ and can be adjusted during training based on gradients computed from the~\texttt{MTL} loss function \( \partial_{\text{MTL}} \).

\subsection{Performance Comparison with \texttt{SOTA} Methods}

In summary, \texttt{DRIFA-Net} achieved a remarkable performance between $95.4\%$ and $99.7\%$ when evaluated  on four diverse medical imaging datasets \texttt{(D1-D4)}. 
In the following, let us delve down deeply in the results. 
The experimental results presented in Table \ref{tab:tab1} indicate that our proposed method surpassed all state-of-the-art multimodal fusion learning approaches, resulting in significant performance enhancements. Specifically, our method shows notable improvements across all metrics used, including accuracy, precision, recall, and F1-score, with gains ranging from $0.2\%$ to $11.4\%$ over other~\texttt{SOTA} models.

Additionally, we applied our \texttt{DRIFA} method to the \texttt{Inception-v3} network \cite{szegedy2016rethinking} and \texttt{SegNet} \cite{badrinarayanan2017segnet} to evaluate its performance when integrated with these architectures on the \texttt{D1–D4} datasets and the \texttt{D5} dataset, respectively. 
The results are shown in Tables~\ref{tab:tab1_2} and~\ref{tab:tab1_3} respectively. 
\texttt{DRIFA-Net} demonstrated significant performance improvements of 1.1\% to 3.2\% over the leading state-of-the-art multimodal fusion learning methods on the \texttt{D1–D4} datasets with \texttt{Inception-v3} network (Table \ref{tab:tab1_2}). It also achieved strong results on the \texttt{D5} dataset, with performance gains ranging from 1.4\% to 5\% compared to the top competitive model -- \texttt{M2}, on \texttt{SegNet} model (Table \ref{tab:tab1_3}) on~\texttt{BraTS2020} dataset. 

\begin{table}[t]
\centering
\caption{Performance comparison of our proposed~\texttt{DRIFA-Net} (\texttt{M5}) with existing multimodal fusion learning approaches (\texttt{M1-M4}) on four benchmark datasets (\texttt{D1-D4}).}
\label{tab:tab1}
\scalebox{0.52}{
\begin{tabular}{cccccccccccc}
\toprule
Dataset & Method & Acc & Prec & Rec & F1 & Dataset & Method & Acc & Prec & Rec & F1 \\
\midrule
\multirow{6}{*}{\texttt{D1}} & M1 & 97.9 & 88.5 & 84.8 & 86.5 & \multirow{6}{*}{\texttt{D2}} & M1 & 95.1 & 95.1 & 95.1 & 95.1 \\
                    & M2 & 97.4 & 95.5 & 97.4 & 96.5 &                      & M2 & 91.9 & 92.5 & 92.1 & 92.3 \\
                    & M3 & 97.2 & 95.9 & 98 & 97.2 &                      & M3 & 91 & 91.5 & 91.5 & 91.5 \\
                    & M4 & 94.4 & 94.2 & 93.8 & 93.8 &                      & M4 & 95.08 & 95.06 & 95.1 & 95.06 \\
                    & \textbf{M5} & \textbf{98.2}  & \textbf{96.4}  & \textbf{99.5}  & \textbf{97.9}  &                      & \textbf{M5} & \textbf{95.6} & \textbf{95.6} & \textbf{95.4} & \textbf{95.5} \\
\midrule
\multirow{6}{*}{\texttt{D3}} & M1 & 98.1 & 98.2 & 96.3 & 97.5 & \multirow{6}{*}{\texttt{D4}} & M1 & 91.5 & 98.8 & 97.8 & 98.3 \\
                    & M2 & 97.9 & 98.0 & 99.8 & 98.0 &                      & M2 & 99.5 & 99.5 & 99.0 & 99.2 \\
                    & M3 & 97.2 & 97.0 & 97.0 & 97.0 &                      & M3 & 98.7 & 97.5 & 97.2 & 97.2 \\
                    & M4 & 97.4 & 98.3 & 96.9 & 97.3 &                      & M4 & 98.2 & 98.3 & 98.2 & 98.1 \\
                    & \textbf{M5} & \textbf{98.4} & \textbf{98.4} & \textbf{98.4} & \textbf{98.4} &                      & \textbf{M5} & \textbf{99.7} & \textbf{99.7} & \textbf{99.3} & \textbf{99.5} \\
\midrule
\end{tabular}}
\vspace{-.4cm}
\end{table}

\begin{table}[t]
\centering
\caption{Performance comparison of our proposed~\texttt{DRIFA} approach integrated with \texttt{Inception-v3} model, named \texttt{DRIFA-Net} (\texttt{M5}) with existing multimodal fusion learning approaches (\texttt{M1-M4}) on four benchmark datasets (\texttt{D1-D4}). 
}
\label{tab:tab1_2}
\scalebox{0.5}{
\begin{tabular}{cccccccccc}
\toprule
Dataset & Method & Acc & F1 & AUC & Dataset & Method & Acc & F1 & AUC \\
\midrule
\multirow{6}{*}{\texttt{D1}} & M1 & 97.9 & 91.5 & 95.4 & \multirow{6}{*}{\texttt{D2}} & M1 & 95.5 & 95.5 & 95.9 \\
                    & M2 & 96.5 & 96.3 & 96.5 &                      & M2 & 92.8 & 92.8 & 93.5  \\
                    & M3 & 97.1 & 94.5 & 96.8 &                      & M3 & 91.7 & 92.3 & 92.8 \\
                    & M4 & 95.3 & 94.7 & 95.8  &                      & M4 & 94.8 & 95.1 & 95.5 \\
                    & \textbf{M5} & \textbf{100}  & \textbf{100}  & \textbf{100}  &                      & \textbf{M5} & \textbf{98.1} & \textbf{98.1} & \textbf{98.9} \\
\midrule
\multirow{6}{*}{\texttt{D3}} & M1 & 96.8 & 96.5 & 96.5 & \multirow{6}{*}{\texttt{D4}} & M1 & 93.7 & 96.4 & 95.2 \\
                    & M2 & 96.8 & 96.05 & 96.5 &                      & M2 & 98.1 & 97.8 & 98.1 \\
                    & M3 & 96.3 & 95.9 & 96.4 &                      & M3 & 98.5 & 98.2 & 98.1 \\
                    & M4 & 96.5 & 96.3 & 96.3 &          & M4 & 97.8 & 97.8 & 98  \\
                    & \textbf{M5} & \textbf{97.9} & \textbf{97.7} & \textbf{98.5} &                      & \textbf{M5} & \textbf{99.7} & \textbf{99.7} & \textbf{99.5}  \\
\midrule
\end{tabular}}
\vspace{-.4cm}
\end{table}

\begin{table}[t]
    \centering
    \caption{Performance comparison of our proposed~\texttt{DRIFA} approach is integrated with \texttt{SegNet} model, named \texttt{DRIFA-Net} (\texttt{M5}) with existing multimodal fusion learning approaches (\texttt{M1-M4}) on \texttt{BraTS2020} dataset.}
    \label{tab:tab1_3}
    \scalebox{0.5}{ 
    \begin{tabular}{c|cccc|cccc}
        \hline
        \multirow{2}{*}{Method} & \multicolumn{4}{c}{Dice Score (\%)} & \multicolumn{4}{|c}{Average Performance} \\
        \cline{2-9} 
        & WT & TC & ET & Average & Acc & AUC & SN & SP \\
        \hline
        M1 & 85.3 & 79.8 & 75.6 & 80.2 & 86.2 & 84.4 & 80.6 & 89.7 \\
        M2 & 89.9 & 85.1 & 79.8 & 84.9 & 90.5 & 86.9 & 84.8 & 92.5 \\
        M3 & 85.7 & 80.2 & 77.1 & 81 & 86.8 & 84.9 & 81.1 & 90.5 \\
        M4 & 84.5 & 78.5 & 75.2 & 79.4 & 85.9 & 82.8 & 79.9 & 88.02 \\
        M5 & \textbf{93.6} & \textbf{90.5} & \textbf{85.6} & \textbf{89.9} & \textbf{93.6} & \textbf{90.14} & \textbf{88.6} & \textbf{93.9} \\
        \hline
    \end{tabular}}
\vspace{-.4cm}
\end{table}


As we discussed earlier, our method excels over existing approaches due to their limited applicability across diverse medical imaging modalities, such as dermoscopy, pap smear cell images, \texttt{MRI}, and \texttt{CT-scan}. These methods often fail to harness the benefits of multiple domains, restricting their ability to capture robust multimodal information and thereby improve model performance. 
For the \texttt{D5} dataset, the existing \texttt{M2}  approach performs well but still achieves limited results due to its lack of focus on leveraging multiple attention methods for enhanced representation learning.
Our approach addresses these limitations by integrating multiple attention methods (\texttt{MFA} and \texttt{MIFA}) to enhance modality-specific and multimodal representation learning, achieving significant performance gains.

\begin{table}[t]
\centering
\caption{
Comparative evaluation of \texttt{MFA} (multi-branch fusion attention) and \texttt{MIFA} (multimodal information fusion attention) components in~\texttt{DRIFA} -- showcasing significant performance enhancement on two benchmark datasets \texttt{D1} and \texttt{D2}. }
\label{tab:tab2}
\scalebox{0.5}{
\begin{tabular}{cccccccc}
\toprule
Dataset & \texttt{MFA} & \texttt{MIFA} & Method & Acc & Prec & Rec & F1 \\
\midrule
 \multirow{4}{*}{\texttt{D1}} & x & x  & DRIFA-Net (Baseline) & 94.85 & 94.4 & 95.2 & 94.7 \\
 & $\checkmark$ & x & DRIFA-Net + MFA & 95.9 & 95.8 & 96.5 & 96.2 \\ 
 & x & $\checkmark$ & DRIFA-Net + MIFA & 96.3 & 96.3 & 96.8 & 96.6 \\ 
 & $\checkmark$ & $\checkmark$ & \textbf{DRIFA-Net + MFA + MIFA} & \textbf{98.2} & \textbf{96.4} & \textbf{99.5} & \textbf{97.9} \\ \hline 

 \multirow{4}{*}{\texttt{D2}}  & x & x & DRIFA-Net & 90.1 & 88.7 & 89.2 & 88.9 \\  
 & $\checkmark$ & x & DRIFA-Net + MFA & 93.8 & 93.6 & 93.8 & 93.8 \\ 
 & x & $\checkmark$ & DRIFA-Net + MIFA & 92.7 & 93.5 & 94.7 & 94.2 \\ 
 & $\checkmark$ & $\checkmark$ & \textbf{DRIFA-Net + MFA + MIFA} & \textbf{95.6} & \textbf{95.6} & \textbf{95.4} & \textbf{95.5} \\ \hline
\end{tabular}}
\vspace{-.3cm}
\end{table}
\begin{table}[t]
\centering
\caption{Experimental evaluation of each \texttt{DRIFA} component—\texttt{HIFA}, \texttt{CLIA}, \texttt{MGIFA}, and \texttt{MLIFA}—on performance by adding and removing them, using datasets \texttt{D1} and \texttt{D2}.}

\label{tab:tab3}
\scalebox{0.6}{
\begin{tabular}{ccccccccc}
\hline
Dataset & \texttt{HIFA} & \texttt{CLIA} & \texttt{MGIFA} & \texttt{MLIFA} & Accuracy & Precision & Recall & F1 score \\ \hline
\multirow{4}{*}{\texttt{D1}} & x & x & x & x & 94.8 & 94.4 & 95.2 & 94.7 \\ 
 & $\checkmark$ & x & $\checkmark$ & x & 95.7 & 95.7 & 95.9 & 95.7 \\ 
 & x & $\checkmark$ & x & $\checkmark$ & 95.1 & 95.1 & 94.9 & 94.9 \\ 
 & $\checkmark$ & $\checkmark$ & $\checkmark$ & $\checkmark$ & \textbf{98.2} & \textbf{96.4} & \textbf{99.5} & \textbf{97.9} \\ \hline
\multirow{4}{*}{\texttt{D2}} & x & x & x & x & 90.1 & 88.7 & 89.2 & 88.9 \\ 
 & $\checkmark$ & x & $\checkmark$ & x & 93.2 & 93.2 & 93.5 & 93.2 \\ 
 & x & $\checkmark$ & x & $\checkmark$ & 91.9 & 91.9 & 92.2 & 92 \\ 
 & $\checkmark$ & $\checkmark$ & $\checkmark$ & $\checkmark$ & \textbf{95.6} & \textbf{95.6} & \textbf{95.4} & \textbf{95.5} \\ \hline
\end{tabular}}
\vspace{-.3cm}
\end{table}

\begin{table}[t]
\centering
\caption{Experimental evaluation on the impact of parameters \texttt{$\omega_d$}, \texttt{$\omega_l$}, \texttt{$\omega_c$} for \texttt{MFA} (Eq.~\ref{eq:eq5}), and \texttt{$\omega_{d_m}$}, \texttt{$\omega_{l_m}$}, \texttt{$\omega_{c_m}$} for~\texttt{MIFA} (Eq.~\ref{eq:mifa_modulation}) with benchmark dataset \texttt{D1}.}
\label{tab:tab6}
\scalebox{0.58}{
\begin{tabular}{cccccc|cccccc}
\hline

 $\omega_d$  & $\omega_l$  & $\omega_c$  & Method                & Acc   & F1    & $\omega_{d_m}$  & $\omega_{l_m}$  & $\omega_{c_m}$  & Method                & Acc   & F1    \\ \hline
         \ding{55} & \ding{55} & \ding{55} &          & 94.85 & 94.7  & \ding{55}   & \ding{55}   & \ding{55}   &         & 94.85 & 94.7  \\ 
         \ding{55} & \ding{55} & \ding{51} &                     & 95.21 & 95.18 & \ding{55}   & \ding{55}   & \ding{51}  &                      & 95.30 & 95.4  \\ 
         \ding{55} & \ding{51} & \ding{51} &    \texttt{DRIFA-}                 & 95.47 & 95.62 & \ding{55}   & \ding{51}  & \ding{51}  & \texttt{DRIFA-}                     & 95.85 & 96.0  \\ 
         \ding{51} & \ding{55} & \ding{51} &  \texttt{Net+MFA}                    & 95.80 & 96.05 & \ding{51}  & \ding{55}   & \ding{51}  & \texttt{Net+MIFA}                   & 95.99 & 96.2  \\ 
         \ding{51} & \ding{51} & \ding{55} &                      & 95.57 & 95.83 & \ding{51}  & \ding{51}  & \ding{55}   &                      & 96.05 & 96.43 \\ 
         \ding{51} & \ding{51} & \ding{51} &                      & \textbf{95.90} & \textbf{96.20} & \ding{51}  & \ding{51}  & \ding{51}  &                   &    \textbf{96.3} & \textbf{96.6} 
\\ \hline
\end{tabular}}
\vspace{-.4cm}
\end{table}

\begin{figure*}[t]
    \centering
    \includegraphics[width=0.6\textwidth,
    height=0.163\textheight]{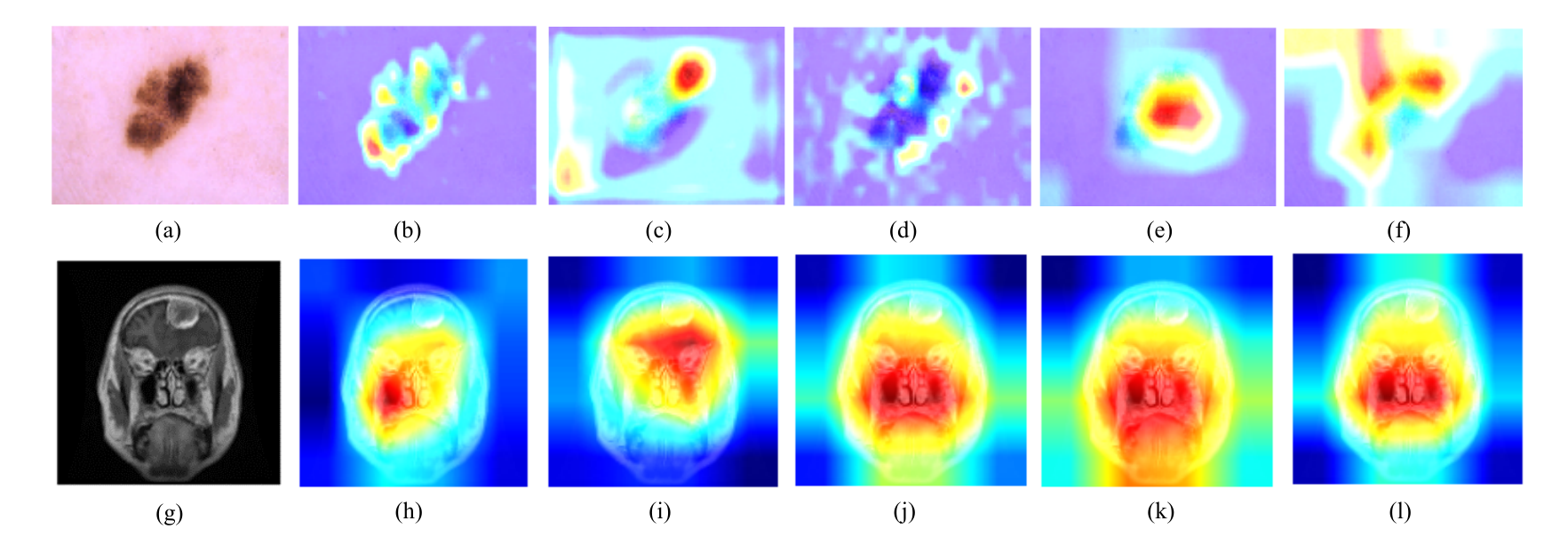}
    \vspace{-.2cm}
    \caption{Visual representation of the important regions highlighted by our proposed \texttt{DRIFA-Net} and four~\texttt{SOTA} methods using the \texttt{GRAD-CAM} technique on two benchmark datasets \texttt{D1} and \texttt{D3}. \textbf{(a)} and \textbf{(g)} display the original images, while \textbf{(b)} and \textbf{(h)} present results for \texttt{Gloria}, \textbf{(c)} and \textbf{(i)} for \texttt{MTF with MA}, \textbf{(d)} and \textbf{(j)} for \texttt{CAF}, \textbf{(e)} and \textbf{(k)} for \texttt{MTTU-Net}, and \textbf{(f)} and \textbf{(l)} for our proposed \texttt{DRIFA-Net}.}
    \label{fig:fig5}
    \vspace{-.2cm}
\end{figure*}
\begin{figure*}[t]
    \centering
    \includegraphics[width=0.95\textwidth, height=0.2\textheight]{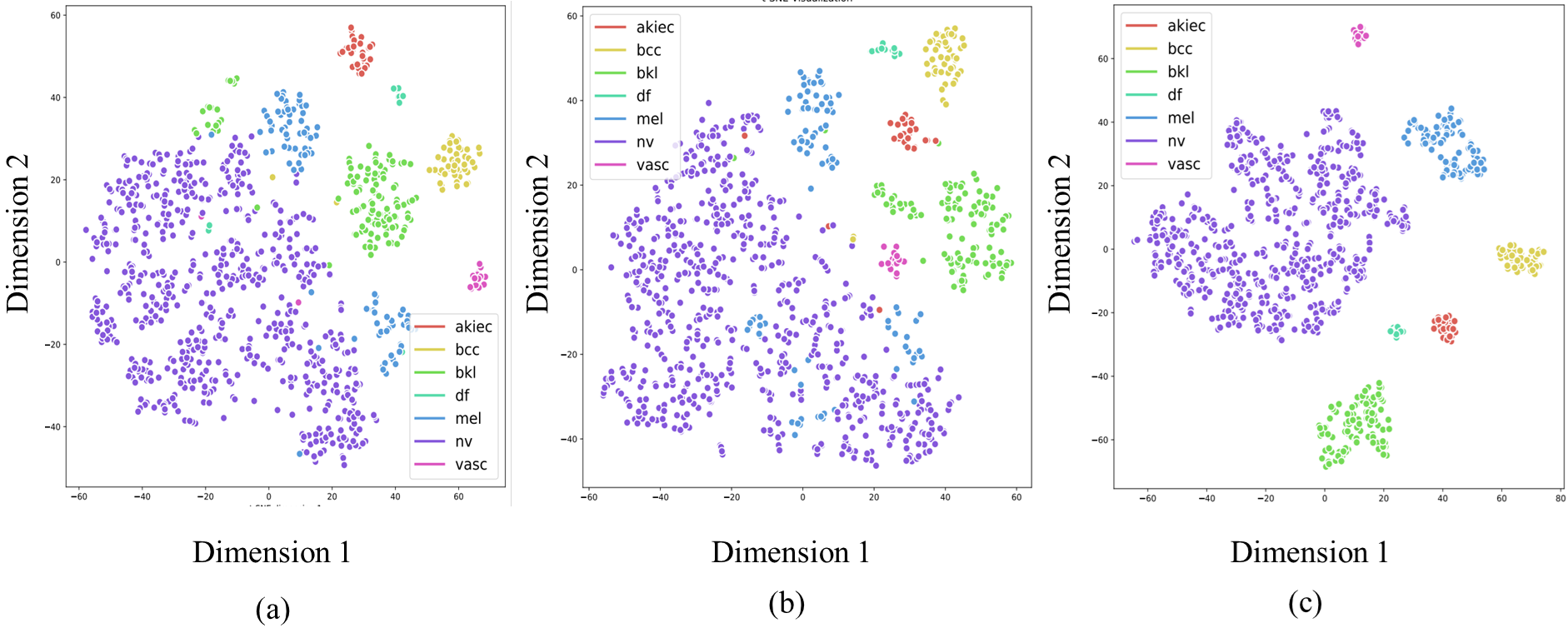}
    \vspace{-.2cm}
    \caption{\texttt{T-SNE} visualization of different models applied to the dermoscopy images of the \texttt{D1} dataset, where \textbf{(a)} represents the \texttt{T-SNE} visualization of \texttt{Gloria}, \textbf{(b)} of \texttt{MTTU-Net}, and \textbf{(c)} of our proposed~\texttt{DRIFA-Net}.}
    \vspace{-.25cm}
    \label{fig:fig6}
\end{figure*}

\subsection{Ablation Study}

We evaluated the impact of different components of our proposed~\texttt{DRIFA} method on two benchmark datasets~\texttt{D1} and~\texttt{D2}~\cite{tschandl2018ham10000, plissiti2018sipakmed} -- focusing on~\texttt{MFA} and~\texttt{MIFA} modules.
The results are given in Table \ref{tab:tab2}. 
It can be seen that~\texttt{DRIFA} approach incorporating both modules outperformed versions employing only one component, achieving performance enhancements ranging from $0.5\%$ to $10\%$. 
This demonstrated the effectiveness and efficacy of our proposed modules.
Again as discussed earlier, this superiority stems from leveraging distinct attention mechanisms for enhancing modality-specific representations and improving multi-modal shared representations simultaneously.

We assessed the impact of each component within the attention modules, specifically, the \texttt{HIFA} and \texttt{CLIA} components of the multi-branch fusion attention module, and the \texttt{MGIFA} and \texttt{MLIFA} components of the multimodal information fusion attention module. 
These evaluations are done on two benchmark datasets \texttt{D1} and \texttt{D2}, and the results are shown in Table \ref{tab:tab3}. 
The experimental results demonstrate that our proposed~\texttt{DRIFA} approach, incorporating all attention components, significantly outperforms versions utilizing at most one component to at least one component, with performance improvements ranging from $0.7\%$ to $5.5\%$. 
One can infer that the limited performance of approaches using only one component arises due to their focus on either enhancing single-modal or multimodal local information, or diverse multimodal global information from modality-specific input features. 
In contrast, (as discussed earlier) our proposed~\texttt{DRIFA} approach designs one attention mechanism for modality-specific representation learning and another for improving multimodal shared representation learning. 
This strategy enhances multimodal representation learning across both global and local contexts in the input data, leading to superior performance.
\begin{table}[t]
\centering
\vspace{-.2cm}
\caption{Uncertainty quantification of~\texttt{DRIFA-Net} predictions on two benchmark datasets \texttt{D1} and \texttt{D2}. \texttt{DRIFA-Net + UQ} denotes~\texttt{DRIFA} with uncertainity quantification.}
\label{tab:tab4}
\scalebox{0.55}{
\begin{tabular}{cccccccccccc}
\toprule
Dataset & Method & Accuracy & Precision & Recall & F1  \\
\hline

\texttt{D1} & \texttt{DRIFA-Net + UQ} & 97.5 & 95.6 & 98.9 & 97.3 \\ \texttt{D2} & \texttt{DRIFA-Net + UQ} & 95.1 & 95.2 & 94.9 & 95.05\\ 
\texttt{D1} &\texttt{DRIFA-Net} & 98.2 & 96.4 & 99.5 & 97.9 \\ \texttt{D2} & \texttt{DRIFA-Net} & 95.6 & 95.6 & 95.4 & 95.5 \\

\hline
\end{tabular}}
\vspace{-.6cm}
\end{table}

Finally, we evaluate the performance of~\texttt{DRIFA-Net + MFA} on D1 dataset with the inclusion and exclusion of learnable parameters $\omega_d, \omega_l, \omega_c$, as discussed in Eq.~\ref{eq:eq5} in Section~\ref{subsubsec_RRA}, and shown in Table~\ref{tab:tab6}. 
We also compare the performance of~\texttt{DRIFA-Net + MIFA} on D1 dataset with the inclusion and exclusion of  $\omega_{d_m}, \omega_{l_m}, \omega_{c_m}$, discussed in Eq.~\ref{eq:mifa_modulation} in Section~\ref{subsubsec_MIFA} and shown in Table~\ref{tab:tab6}.
It can be seen that the best performance is achieved on D1 dataset when the three components are included highlighting their efficacy and important role towards the working of~\texttt{DRIFA-Net + MFA} and~\texttt{DRIFA-Net + MIFA}.

\subsection{Impact of uncertainty quantification}

We utilized an ensemble Monte Carlo dropout (\texttt{MCD}) strategy to assess the prediction uncertainty of our proposed approach across each medical imaging modality, as detailed in Table \ref{tab:tab4}. 
\texttt{DRIFA} with uncertainty quantification is denoted as~\texttt{DRIFA-Net + UQ} and led to a marginal drops of $0.4\%$ to $0.7\%$ on the two benchmark datasets when compared to our model without uncertainty estimation.
While \texttt{UQ} can be computationally intensive, it was chosen for its reliability and flexibility. It handles various models without strong assumptions, providing a comprehensive view of outcomes and robust uncertainty estimation, crucial for accurate clinical decision-making.

\subsection{Impact of qualitative analysis}
We performed a qualitative analysis using \texttt{Grad-CAM} to assess the efficacy of our proposed method. This visualization highlighted regions of highest importance in the two benchmark datasets \texttt{D1} and \texttt{D3}, as shown in Fig.~\ref{fig:fig5}. Additionally, the \texttt{T-SNE} plot for the \texttt{D1} dataset in Fig.~\ref{fig:fig6} further validated our model's decisions by emphasizing areas crucial to the prediction scores.

\section{Conclusion} \label{sec:concl}

In this paper, we proposed a dual information fusion attention approach to enhance multimodal fusion learning, making it applicable to diverse disease classification tasks across medical imaging modalities such as cervical, skin, lung cancer, and brain tumors. By combining multi-branch fusion attention and multimodal information attention modules, we outperform existing state-of-the-art methods shown through extensive experiments. 
Future work will focus on expanding our approach for more medical imaging modalities and optimizing computational efficiency.

{\small
\bibliographystyle{ieee_fullname}
\bibliography{egbib}

\begin{thebibliography}{10}\itemsep=-1pt

\bibitem{Abdelhalim2021}
I.~S.~A. Abdelhalim, M.~F. Mohamed, and Y.~B. Mahdy.
\newblock Data augmentation for skin lesion using self-attention based progressive generative adversarial network.
\newblock {\em Expert Systems with Applications}, 165:113922, 2021.

\bibitem{alyasriy2020iq}
Hamdalla Alyasriy and A Muayed.
\newblock The iq-othnccd lung cancer dataset.
\newblock {\em Mendeley Data}, 1(1):1--13, 2020.

\bibitem{alzahrani2023convattenmixer}
Salha~M. Alzahrani.
\newblock Convattenmixer: Brain tumor detection and type classification using convolutional mixer with external and self-attention mechanisms.
\newblock {\em Journal of King Saud University-Computer and Information Sciences}, 35(10):101810, 2023.

\bibitem{Anand2023}
V. Anand, S. Gupta, D. Koundal, and K. Singh.
\newblock Fusion of u-net and cnn model for segmentation and classification of skin lesion from dermoscopy images.
\newblock {\em Expert Systems with Applications}, 213:119230, 2023.

\bibitem{badrinarayanan2017segnet}
Vijay Badrinarayanan, Alex Kendall, and Roberto Cipolla.
\newblock Segnet: A deep convolutional encoder-decoder architecture for image segmentation.
\newblock {\em IEEE transactions on pattern analysis and machine intelligence}, 39(12):2481--2495, 2017.

\bibitem{Cai2023}
G. Cai, Y. Zhu, Y. Wu, X. Jiang, J. Ye, and D. Yang.
\newblock A multimodal transformer to fuse images and metadata for skin disease classification.
\newblock {\em The Visual Computer}, 39(7):2781--2793, 2023.

\bibitem{celik2024development}
Muhammed Celik and Ozkan Inik.
\newblock Development of hybrid models based on deep learning and optimized machine learning algorithms for brain tumor multi-classification.
\newblock {\em Expert Systems with Applications}, 238:122159, 2024.

\bibitem{Chen2023JCRCO}
Q. Chen, M. Li, C. Chen, P. Zhou, X. Lv, and C. Chen.
\newblock Mdfnet: application of multimodal fusion method based on skin image and clinical data to skin cancer classification.
\newblock {\em Journal of Cancer Research and Clinical Oncology}, 149(7):3287--3299, 2023.

\bibitem{cheng2022fully}
J. Cheng, J. Liu, H. Kuang, and J. Wang.
\newblock A fully automated multimodal mri-based multi-task learning for glioma segmentation and idh genotyping.
\newblock {\em IEEE Transactions on Medical Imaging}, 41(6):1520--1532, June 2022.

\bibitem{Dhar2021a}
J. Dhar.
\newblock An adaptive intelligent diagnostic system to predict early stage of parkinson's disease using two-stage dimension reduction with genetically optimized lightgbm algorithm.
\newblock {\em Neural Computing and Applications}, 34(6):4567--4593, 2021.

\bibitem{georgescu2023multimodal}
M.~I. Georgescu, R.~T. Ionescu, A.~I. Miron, O. Savencu, N.~C. Ristea, N. Verga, and F.~S. Khan.
\newblock Multimodal multi-head convolutional attention with various kernel sizes for medical image super-resolution.
\newblock In {\em Proceedings of the IEEE/CVF winter conference on applications of computer vision}, pages 2195--2205, 2023.

\bibitem{Han2024}
Q. Han, X. Qian, H. Xu, K. Wu, L. Meng, Z. Qiu, and X. Gao.
\newblock Dm-cnn: Dynamic multiscale convolutional neural network with uncertainty quantification for medical image classification.
\newblock {\em Computers in Biology and Medicine}, 168:107758, 2024.

\bibitem{He2023}
X. He, Y. Wang, S. Zhao, and X. Chen.
\newblock Co-attention fusion network for multimodal skin cancer diagnosis.
\newblock {\em Pattern Recognition}, 133:108990, 2023.

\bibitem{Hemalatha2023}
K. Hemalatha, V. Vetriselvi, and M. Dhandapani.
\newblock Cervixfuzzyfusion for cervical cancer cell image classification.
\newblock {\em Biomedical Signal Processing and Control}, 85:104920, 2023.

\bibitem{huang2021gloria}
S.~C. Huang, L. Shen, M.~P. Lungren, and S. Yeung.
\newblock Gloria: A multimodal global-local representation learning framework for label-efficient medical image recognition.
\newblock In {\em Proceedings of the IEEE/CVF International Conference on Computer Vision}, pages 3942--3951, 2021.

\bibitem{islam2022mumu}
M.~M. Islam and T. Iqbal.
\newblock Mumu: Cooperative multitask learning-based guided multimodal fusion.
\newblock In {\em Proceedings of the AAAI Conference on Artificial Intelligence}, volume~36, pages 1043--1051, June 2022.

\bibitem{kawahara2018seven}
Jeremy Kawahara, Seyed Daneshvar, Giuseppe Argenziano, and Ghassan Hamarneh.
\newblock Seven-point checklist and skin lesion classification using multitask multimodal neural nets.
\newblock {\em IEEE journal of biomedical and health informatics}, 23(2):538--546, 2018.

\bibitem{kihara2022policy}
Y. Kihara, G. Montesano, A. Chen, N. Amerasinghe, C. Dimitriou, A. Jacob, and A.~Y. Lee.
\newblock Policy-driven, multimodal deep learning for predicting visual fields from the optic disc and oct imaging.
\newblock {\em Ophthalmology}, 129(7):781--791, 2022.

\bibitem{Kim2023}
S. Kim, T.~G. Purdie, and C. McIntosh.
\newblock Cross-task attention network: Improving multitask learning for medical imaging applications.
\newblock In {\em International Conference on Medical Image Computing and Computer-Assisted Intervention}, pages 119--128. Springer Nature Switzerland, 2023.

\bibitem{li2021multimodal}
Y. Li, J. Zhao, Z. Lv, and Z. Pan.
\newblock Multimodal medical supervised image fusion method by cnn.
\newblock {\em Frontiers in Neuroscience}, 15:638976, 2021.

\bibitem{Ling2023}
Y. Ling, Y. Wang, W. Dai, J. Yu, P. Liang, and D. Kong.
\newblock Mtanet: Multitask attention network for automatic medical image segmentation and classification.
\newblock {\em IEEE Transactions on Medical Imaging}, 2023.

\bibitem{Liu2022}
W. Liu, C. Li, M.~M. Rahaman, T. Jiang, H. Sun, X. Wu, and M. Grzegorzek.
\newblock Is the aspect ratio of cells important in deep learning? a robust comparison of deep learning methods for multiscale cytopathology cell image classification: From convolutional neural networks to visual transformers.
\newblock {\em Computers in biology and medicine}, 141:105026, 2022.

\bibitem{Manna2021}
A. Manna, R. Kundu, D. Kaplun, A. Sinitca, and R. Sarkar.
\newblock A fuzzy rank-based ensemble of cnn models for classification of cervical cytology.
\newblock {\em Scientific Reports}, 11(1):14538, 2021.

\bibitem{menze2014multimodal}
Bjoern~H Menze, Andras Jakab, Stefan Bauer, Jayashree Kalpathy-Cramer, Keyvan Farahani, Justin Kirby, Yuliya Burren, Nicole Porz, Johannes Slotboom, Roland Wiest, et~al.
\newblock The multimodal brain tumor image segmentation benchmark (brats).
\newblock {\em IEEE transactions on medical imaging}, 34(10):1993--2024, 2014.

\bibitem{Naveed2024}
A. Naveed, S.~S. Naqvi, T.~M. Khan, and I. Razzak.
\newblock Pca: Progressive class-wise attention for skin lesions diagnosis.
\newblock {\em Engineering Applications of Artificial Intelligence}, 127:107417, 2024.

\bibitem{nickparvar2021}
Msoud Nickparvar.
\newblock Brain tumor {MRI} dataset.
\newblock Data set, 2021.
\newblock Accessed on 3rd March.

\bibitem{Omeroglu2023}
A.~N. Omeroglu, H.~M. Mohammed, E.~A. Oral, and S. Aydin.
\newblock A novel soft attention-based multimodal deep learning framework for multi-label skin lesion classification.
\newblock {\em Engineering Applications of Artificial Intelligence}, 120:105897, 2023.

\bibitem{Pacal2023}
I. Pacal and S. K?l?carslan.
\newblock Deep learning-based approaches for robust classification of cervical cancer.
\newblock {\em Neural Computing and Applications}, 35(25):18813--18828, 2023.

\bibitem{Pedro2022}
R. Pedro and A.~L. Oliveira.
\newblock Assessing the impact of attention and self-attention mechanisms on the classification of skin lesions.
\newblock In {\em 2022 International Joint Conference on Neural Networks (IJCNN)}, pages 1--8. IEEE, 2022.

\bibitem{plissiti2018sipakmed}
Maria~E Plissiti, Panagiotis Dimitrakopoulos, Giorgos Sfikas, Christophoros Nikou, Orestis Krikoni, and Avraam Charchanti.
\newblock Sipakmed: A new dataset for feature and image based classification of normal and pathological cervical cells in pap smear images.
\newblock In {\em 2018 25th IEEE International Conference on Image Processing (ICIP)}, pages 3144--3148. IEEE, October 2018.

\bibitem{Qian2022}
S. Qian, K. Ren, W. Zhang, and H. Ning.
\newblock Skin lesion classification using cnns with grouping of multiscale attention and class-specific loss weighting.
\newblock {\em Computer Methods and Programs in Biomedicine}, 226:107166, 2022.

\bibitem{Song2024}
Y. Song, J. Zou, K.~S. Choi, B. Lei, and J. Qin.
\newblock Cell classification with worse-case boosting for intelligent cervical cancer screening.
\newblock {\em Medical Image Analysis}, 91:103014, 2024.

\bibitem{steyaert2023multimodal}
S. Steyaert, M. Pizurica, D. Nagaraj, P. Khandelwal, T. Hernandez-Boussard, A.~J. Gentles, and O. Gevaert.
\newblock Multimodal data fusion for cancer biomarker discovery with deep learning.
\newblock {\em Nature Machine Intelligence}, 5(4):351--362, 2023.

\bibitem{szegedy2016rethinking}
Christian Szegedy, Vincent Vanhoucke, Sergey Ioffe, Jon Shlens, and Zbigniew Wojna.
\newblock Rethinking the inception architecture for computer vision.
\newblock In {\em Proceedings of the IEEE conference on computer vision and pattern recognition}, pages 2818--2826, 2016.

\bibitem{tabarestani2020distributed}
S. Tabarestani, M. Aghili, M. Eslami, M. Cabrerizo, A. Barreto, N. Rishe, and M. Adjouadi.
\newblock A distributed multitask multimodal approach for the prediction of alzheimer's disease in a longitudinal study.
\newblock {\em NeuroImage}, 206:116317, 2020.

\bibitem{tan2022multi}
K. Tan, W. Huang, X. Liu, J. Hu, and S. Dong.
\newblock A multi-modal fusion framework based on multi-task correlation learning for cancer prognosis prediction.
\newblock {\em Artificial Intelligence in Medicine}, 126:102260, 2022.

\bibitem{tschandl2018ham10000}
Philipp Tschandl, Cliff Rosendahl, and Harald Kittler.
\newblock The ham10000 dataset, a large collection of multi-source dermatoscopic images of common pigmented skin lesions.
\newblock {\em Scientific data}, 5(1):1--9, 2018.

\bibitem{vaswani2017attention}
A Vaswani.
\newblock Attention is all you need.
\newblock {\em Advances in Neural Information Processing Systems}, 2017.

\bibitem{Yang2023}
L. Yang, C. Fan, H. Lin, and Y. Qiu.
\newblock Rema-net: An efficient multi-attention convolutional neural network for rapid skin lesion segmentation.
\newblock {\em Computers in Biology and Medicine}, 159:106952, 2023.

\bibitem{Yutra2023}
A.~Z. Yutra, J. Zheng, X. Li, and A. Endris.
\newblock Skinaacn: An efficient skin lesion classification based on attention augmented convnext with hybrid loss function.
\newblock In {\em Proceedings of the 2023 7th International Conference on Computer Science and Artificial Intelligence}, pages 295--300, 2023.

\bibitem{Zhou2021}
L. Zhou and Y. Luo.
\newblock Deep features fusion with mutual attention transformer for skin lesion diagnosis.
\newblock In {\em 2021 IEEE international conference on image processing (ICIP)}, pages 3797--3801. IEEE, 2021.

\end{thebibliography}
}

\end{document}